\documentclass{article}

\usepackage{microtype}
\usepackage{graphicx}
\usepackage{subfigure}
\usepackage{booktabs} 
\usepackage{tablefootnote}
\usepackage{longtable}
\usepackage{multirow}

\usepackage{amsmath}
\usepackage{amsfonts}
\usepackage[dvipsnames]{xcolor}

\usepackage{hyperref}

\usepackage[accepted]{icml2020}

\begin{document}

\twocolumn[
\icmltitle{PENNI: Pruned Kernel Sharing for Efficient CNN Inference}



\icmlsetsymbol{equal}{*}

\begin{icmlauthorlist}
\icmlauthor{Shiyu Li}{du}
\icmlauthor{Edward Hanson}{du}
\icmlauthor{Hai Li}{du}
\icmlauthor{Yiran Chen}{du}
\end{icmlauthorlist}

\icmlaffiliation{du}{Department of Electrical and Computer Engineering, Duke University, Durham NC, United States}

\icmlcorrespondingauthor{Shiyu Li}{shiyu.li@duke.edu}

\icmlkeywords{Network Pruning, Model Compression and Acceleration, CNN}

\vskip 0.3in
]



\printAffiliationsAndNotice{}  


\begin{abstract}
Although state-of-the-art (SOTA) CNNs achieve outstanding performance on various tasks, their high computation demand and massive number of parameters make it difficult to deploy these SOTA CNNs onto resource-constrained devices. Previous works on CNN acceleration utilize low-rank approximation of the original convolution layers to reduce computation cost. However, these methods are very difficult to conduct upon sparse models, which limits execution speedup since redundancies within the CNN model are not fully exploited. We argue that kernel granularity decomposition can be conducted with low-rank assumption while exploiting the redundancy within the remaining compact coefficients. Based on this observation, we propose PENNI, a CNN model compression framework that is able to achieve model compactness and hardware efficiency simultaneously by (1) implementing kernel sharing in convolution layers via a small number of basis kernels and (2) alternately adjusting bases and coefficients with sparse constraints. Experiments show that we can prune 97\% parameters and 92\%  FLOPs on ResNet18 CIFAR10 with no accuracy loss, and achieve 44\% reduction in run-time memory consumption and a 53\% reduction in inference latency.



\end{abstract}

\section{Introduction}

One of the greatest strengths of Deep Neural Networks (DNNs), specifically Convolutional Neural Networks (CNNs), is their large design space, which innately heightens flexibility and potential for accuracy. Improving model accuracy conventionally involves increasing its size, given sufficient training data. This increase in size can come in the form of more layers \cite{he2016deep}, more channels per layer \cite{zagoruyko2016wide}, or more branches \cite{szegedy2015going}. A major drawback of na\"ively increasing model size is the substantial computational power and memory bandwidth required to train and run inference tasks. To address this issue, multiple methods have been introduced to compress CNN models and increase sparsity \cite{han2015deep, wen2016learning}. Model compression can come in the form of weight quantization \cite{ullrich2017soft} or Low Rank Approximation (LRA) \cite{denton2014exploiting}. 

LRA utilizes matrix factorization to decompose weight matrices into the product of two low rank matrices, thus reducing computation cost. Some works \cite{lebedev2014speeding, tai2016convolutional} use tensor decomposition to represent the original weight with the outer product of one-dimensional tensors (i.e., vectors) . The speedup and parameter reduction of these methods are notable; however, current approaches are limited because they do not consider redundancies in CNN parameters.


Model sparsity can be induced via various pruning techniques, most of which are categorized under \textit{structured} or \textit{unstructured}. On one hand, unstructured pruning aims to remove unimportant weights of a network, irrespective of its location. By targeting the least important weights in a model, unstructured pruning has minimal impact on overall accuracy while achieving a high sparsity level. However, the undefined distribution of pruned weights makes it challenging to compress the model's representation in memory. On the other hand, structured pruning achieves sparsity by removing entire DNN structures (e.g. filter-channels, filters, layers, etc.) that are deemed unimportant, which may impact a model's performance by inadvertently removing sensitive parameters. Such predictable pruning patterns open the avenue for efficient model storage and computation. It is important to note that merely applying structured pruning is not enough to fully reap hardware efficiency benefits. Without additional changes to the underlying representation in memory or the model's training or inference stage algorithms, conventional DNN platforms still fall victim to inefficient memory transfers and computations.

\comment{; conceptually pruned parameters are merely replaced with zeroes in practice, which imposes useless data transfers and zero-computations on conventional DNN hardware platforms.

Because some parameters in a pruned structure may hold greater influence on a model's performance than those in non-pruned structures, structured pruning typically has greater accuracy degradation \ed{[MAY NEED CITATION]}.

Although various tricks[CITATIONS] are proposed to utilize the sparsity of the weights, a notable gap still lies between the model sparsity and the improvement on the hardware efficiency.

For the purpose of this paper, \ed{"DNN structures"} refers to any logical collection of parameters in a network that is mapped onto a non-scalar tensor of any dimension (i.e., filter-channels, filters, layers, etc.).

The strength of structured pruning lies in its predictable pruning patterns, which effectively eliminates entire dimensions of tensors representing a network's parameters.}


In this paper, we propose Pruned kernel sharing for Efficient CNN Inference (PENNI), a CNN model compression framework that overcomes these challenges by decomposing layer parameters into tiny sets of basis kernels and accompanying coefficient matrices. This method can benefit inference efficiency by organizing the involved coefficients and computation flow in a hardware-friendly manner. High compression rate is achieved by applying $l1$-regularization to the coefficients. The structural redundancies are further explored in a model shrinkage procedure.
We evaluate our method on CIFAR10 and ImageNet with VGG16, ResNet and AlexNet. Results show that we can achieve a 98.3\% reduction on parameters and a 93.3\% reduction on FLOPs with less than 0.4\% accuracy drop. \comment{94.5\% parameters and 76.9\% FLOPs can be eliminated on ResNet56 with 0.2\% accuracy drop. On ImageNet, a 87.3\% reduction on parameters and a 94.7\% reduction on FLOPs can be achieved with 2.4\% accuracy loss.} PENNI outperforms state-of-the-art (SOTA) pruning schemes in addition to being more efficient for hardware implementation. Our code is avaliable at: \url{https://github.com/timlee0212/PENNI}.

Our main contributions are listed as follows:
\begin{itemize}
    \item We propose a hardware-friendly CNN model compression framework, PENNI. We apply filter decomposition to generate a limited set of basis kernels and corresponding coefficient matrix. Sparsity is achieved by applying $l1$-regularization to coefficient matrices in the retraining process. Structural redundancies are then explored via a model shrinking procedure. 
    
    
    \item Hardware inference efficiency is directly benefited through model shrinking with no modifications to inference algorithm. Further speedup can be brought by computation reorganization of convolutional layers. To avoid restoring original filter tensors, we can separate basis kernel convolutions from their weighted sum computation. Keeping the two computation steps distinct opens the avenue for exposing all pruned coefficients, thus leveraging coefficient sparsity and avoiding wasteful zero-computations.
    
    \item Evaluation on CIFAR-10 and ImageNet with various network architectures proves the effectiveness of the proposed method with significant reduction in both FLOPs and number of parameters with negligible accuracy loss.
\end{itemize}



\comment{ (Backgrounds on DNN and the desire for compact DNN design)

(To address this issue, multiple method are introduced to compress the model.[Maybe also add the compact desgin, i.e. MobileNet or sth.])

However, (The shortcoming issue of current method)

In this paper, we proposed XXX.(a brief intro of the method)

Our main contributions are listed as follows:
\begin{itemize}
    \item We proposed XXX, a hardware friendly DNN compression method for efficient inference. 
    \item (How will this method benefit hardware inference)
    \item Our experiments show that significant reduction on both FLOPS and parameters are achieved while the accuracy drop of the network is kept acceptable.
\end{itemize}

}
\section{Related Work}
\label{sec:rw}
Various methods have been proposed to accelerate CNN inference. These methods either exploit redundancies of CNN models to reduce the number of parameters and computations or introduce lightweight model structures for a given task.

\paragraph{Compact Model Design} Previous works aim to develop resource-efficient model structures to reduce computation requirements and improve latency. Lin et al. \yrcite{lin2013network} propose global average pooling and 1x1 convolution, which are widely adopted in the later compact architectures. SqueezeNet \cite{iandola2016squeezenet} utilizes both structures to reduce the number of channels and remove fully-connected layers. A similar idea appears in InceptionNet \cite{szegedy2015going}, while a later version \cite{szegedy2016rethinking} extends the idea by spatially separating the convolutional layers. MobileNet \cite{howard2017mobilenets} uses depthwise separable convolution to reduce the computation cost by splitting the original convolutional layer channel-wise. Its following version, MobileNet V2 \cite{sandler2018mobilenetv2}, adopts residual connections and introduces the inverted bottleneck module to improve efficiency. Xie et al. \yrcite{xie2017aggregated} enhance the expressiveness of the depthwise convolution by allowing limited connectivity within groups, while later ShuffleNet \cite{zhang2018shufflenet} adopts the grouped convolution. In addition to the manually designed compact architectures listed above, Neural Architecture Search (NAS) methods aim to automatically find architectures with optimal balances of compactness and performance. Multiple such works \cite{tan2019mnasnet, cai2018proxylessnas, liu2018darts, tan2019efficientnet} generate architectures that outperform manually designed ones.

\paragraph{Low Rank Approximation}
Low Rank Approximation (LRA) method decomposes the original weights into several low rank matrices \cite{denton2014exploiting,zhang2015accelerating} or low dimension tensors \cite{lebedev2014speeding, kim2015compression}. Denton et al. \yrcite{denton2014exploiting} utilize Singular Value Decomposition (SVD) to conduct the decomposition, whereas Zhang et al. \yrcite{zhang2015accelerating} take nonlinear activations into account to obtain the decomposition while minimizing error of the response. Kim et al. \yrcite{kim2015compression} adopt Tucker Decomposition to compress the kernel tensor. Lebedev et al. \yrcite{lebedev2014speeding} use canonical polyadic (CP) decomposition. In addition, Learning Structured Sparsity \cite{wen2016learning} and Centripetal-SGD \cite{ding2019centripetal}  directly train the DNN with low rank constraints. These tensor decomposition methods rely on the rank selection, which is an ill-posed problem, while the matrix factorization methods have limited speedup since redundancies in the standalone weight values are not considered.

\begin{figure*}[t]
    \centering
    \includegraphics[width=0.9\linewidth]{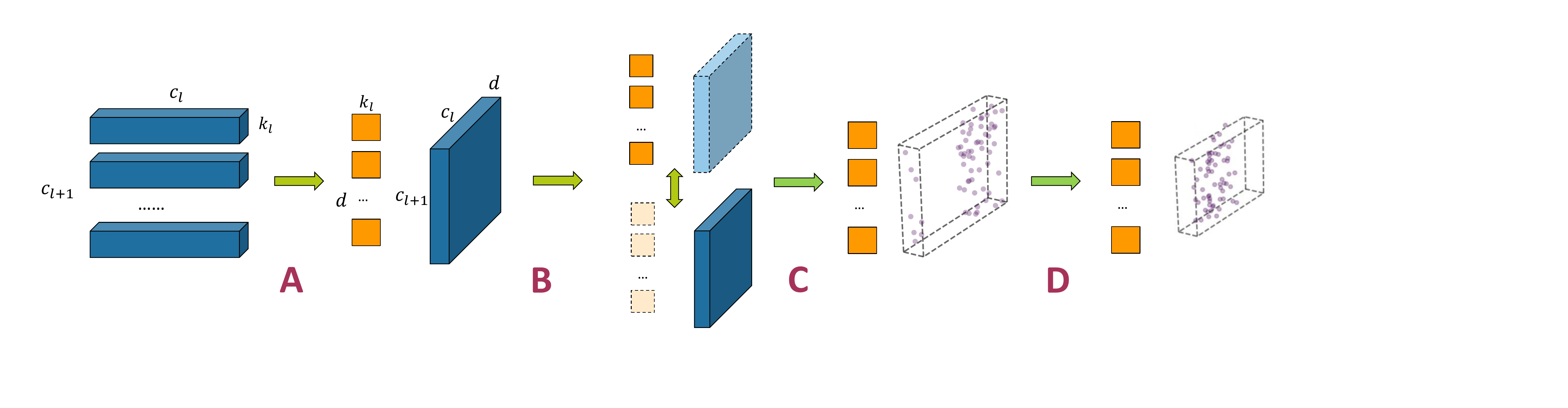}
        \caption{Overview of PENNI. There are four phases in the proposed framework: \textbf{A.} Decompose the filters into $d$-dimension basis and the corresponding coefficient matrix; \textbf{B.} Recover the model performance by alternatively training basis and coefficients with sparsity regularization applied to coefficients; \textbf{C.} Prune coefficient by magnitude; \textbf{D.} Explore the structure redundancies and shrink the model.}
    \label{fig:framework}
\end{figure*}

\paragraph{Model Pruning}
The idea of weight pruning dates back to the last century. Optimal Brain Damage \cite{lecun1990optimal} proposes pruning weight based on their impact on the loss function. A later work, Optimal Brain Surgeon \cite{hassibi1993second} improves this method by replacing the diagonal Hessian Matrix with an approximated full covariance matrix. However, due to the giant size of the modern DNNs, these methods incur unacceptable computation cost. Han et al. \yrcite{han2015deep} propose pruning weights by comparing the magnitude with a threshold, and achieve the optimal result by iterative pruning and fine-tuning. Guo et al. \yrcite{guo2016dynamic} further improve the sparsity level by maintaining a mask instead of directly pruning the redundant weights. Beyond conventional unstructured pruning methods, various structured pruning methodologies have been proposed to ease translation from sparsity to inference speedup. Wen et al. \yrcite{wen2016learning} and Yang et al. 
\yrcite{yang2019deephoyer} apply group regularizer in the training process to obtain structured sparsity. Liu et al. \yrcite{liu2017learning} apply $l1$-regularization to the scaling factors of batch normalization layers to identify insignificant channels. ThiNet \cite{luo2017thinet} utilizes a data-driven method to prune the channel with the smallest impact on the following layer. In recent works \cite{he2018soft, zhang2018learning, he2019filter, ding2019approximated},  different criteria are adopted to rank the importance of the filter. Louizos et al. \yrcite{louizos2017learning} use stochastic gates to apply $l0$-regularization to the filters. NAS methods also incorporate filter pruning \cite{he2018amc, liu2019metapruning}. Although structured pruning can directly benefit the inference efficiency, its pruning granularity limits the compression rate or accuracy of CNN models.


\section{Proposed Method}
\label{sec:met}
\subsection{Overview}
Fig.\ref{fig:framework} presents the overview of PENNI framework. We first decompose each layer's convolution filters into a few basis kernels and a coefficient matrix. Then, we retrain the decomposed network with sparsity regularization applied to coefficient matrix to recover any lost accuracy. Finally, we prune the redundant coefficients based on magnitude and obtain a compact CNN model.

Before the discussion on the method, we first define the notations that will be used in this paper. We denote the parameters of convolutional layer $l$ as $\mathbf{\theta}^{(l)}\in \mathbb{R}^{c_l\times c_{l+1} \times k_w^l \times k_h^l}$, where $c_l$ is the number of the input channels, $c_{l+1}$ is the number of the output channels, and $k_w^{(l)}$ and $k_h^{(l)}$ are the kernel dimensions. Since most CNN architectures implement a square kernel shape, i.e., $ k_w^{(l)} = k_h^{(l)}$, we denote the kernel shape as $k^{(l)}\times k^{(l)}$ for simplicity; the shape of the kernel does not affect this framework. $\Theta = \{\mathbf{\theta}^{(l)}\}$ is the set of all parameters of convolutional layers of a CNN model. We denote the output features of layer $l$ as $S^{(l)}$, and the input features as $I^{(l)}$. $(X, Y)$ represents the data pairs, while $Y$ is the given label or unknown ground-truth. $\hat{Y}$ represents the network model's prediction. With these notations, the $i$-th channel of a layer's output feature map $S^{(l)}$ can be computed by:
\begin{equation}
    S^{(l)}_i =  \sigma^{(l)}\Big(\big(\sum_{j=1}^{c_{l-1}}I^{(l-1)}_j*\theta^{(l)}_{i,j}\big) + b_j^{(l)}\Big),
\end{equation}
where $\theta^{(l)}_{i,j}$ is the $j$-th kernel of the $i$-th filter, $b_j^{(l)}$ is the bias term of the $j$-th filter and the $\sigma^{(l)}$ is the non-linear function of the layer $l$.

\subsection{Filter Decomposition}
The convolution operation dominates computation cost of CNN inference. Irregular data access and compute patterns make it extremely difficult to efficiently map the operation onto parallel hardware and further improve inference efficiency. We address this issue by reducing the number of convolution operations and offloading the irregular computation to a sequential and simple pattern. 

Previous work \cite{zhang2015accelerating} on accelerating CNN inference utilizes a low rank assumption of output feature subspace to represent the original weight matrix with the multiplication of two low rank matrices, thus reducing the computation required. Low rank assumption is reasonable in this case because the number of output features is comparable with the dimension of the feature space. Recent work \cite{ding2019centripetal} indicates that in most CNNs, regularization on convolutional kernels can push the kernels to be alike one another. Based on this observation, we argue that the low rank assumption can also be applied to the subspace that each convolutional kernel lies in. With this assumption, we approximate the original convolutional layer by sharing a tiny set of basis kernels and representing original kernels with coefficients.

Decomposition at a kernel granularity is done to obtain an approximated layer. This process applies to a single layer a time, so the superscript $l$ is omitted for readability. We first reshape the original weight tensor into a 2D matrix $\theta'\in \mathbb{R}^{c_lc_{l+1} \times k_l^2}$; thus, each kernel can be seen as its row vector $w\in\mathbb{R}^{k_l^2}$. Suppose $\mathbf{U}\subset \mathbb{R}^{k_l^2}$ is a subspace with basis $\mathcal{B}=\{u_1, u_2, ..., u_d\}$ where $d\leq k_l^2$. The objective of decomposition process is to find the subspace that minimizes the error between the projected and original vectors, shown in Equation \ref{eq:obj_decom}.
\begin{equation}
\label{eq:obj_decom}
    \min_{\alpha_w\in\mathbb{R}^{d}}\sum_{w\in\theta'}||w- \alpha_w\mathbf{B}^T||^2.
\end{equation}
$\mathbf{B}=[u_1\ u_2\ ...\ u_d]$ is the basis matrix where each column vector is a basis of the subspace and $\alpha_w$ is the coefficient vector corresponding to the row vector $w$. With this decomposition, the output of each layer is computed by:
\begin{equation}
\label{eq:decom_conv}
    S^{(l)}_i =  \sigma^{(l)}\Big(\big(\sum_{j=1}^{c_{l-1}}I^{(l-1)}_j*(\alpha^{(l)}_{i,j}\mathbf{B}^{(l)T}\big) + b_j^{(l)}\Big),
\end{equation}
where $\alpha^{(l)}_{i,j}$ is the row vector in the coefficient matrix corresponding to the $j$-th kernel of the $i$-th filter. 

The decomposition problem can be formulated as best approximation and is perfectly solved using singular value decomposition (SVD). We first obtain $\Bar{\mathbf{\theta'}}$ by subtracting each row vector with the mean vector, and then compute the covariance matrix $\mathbf{W} = \mathbf{\theta'}^T\mathbf{\theta'}$. Conducting SVD on $\mathbf{W}$, and organizing the singular value by their magnitude, we'll have:
\begin{equation}
    \mathbf{W} = \mathbf{U}\mathbf{\Sigma} \mathbf{V}^T.
\end{equation}
The basis matrix $\mathbf{B}$ is then derived by selecting the first $d$ columns from matrix $\mathbf{U}$ and obtaining the corresponding coefficients by multiplying the $\theta'$ by the projection matrix $\mathbf{BB}^T$. Normally, $k_l^2 << c_lc_{l+1}$ and parameter matrices of a pretrained model are dense, so $W$ is a full rank matrix with the rank $k_l^2$. Thus, the low rank approximation on kernel space makes the SVD computation faster than conducting decomposition at the filter granularity. A singular value may represent the portion of the basis vector contributing to the original vectors; but, rather than selecting $d$ based on it, we leave it as a hyper-parameter providing a trade-off between computational cost and model accuracy. 

\subsection{Retraining}

Although the discussed filter decomposition scheme gives the best approximation of the original parameters in low-rank subspace, the model accuracy may greatly degrade due to varying sensitivity of affected weights. Zhang et al. \yrcite{zhang2015accelerating} address this issue by considering the non-linear block and minimizing response of the layer. However, innate redundancies in the models are not exploited, which limits the compression rate and the speedup. Thus, we incorporate a retraining process for twofold benefits: recover the model accuracy and exploit redundancy within the CNN structure through coefficient sparsity regularization. The objective of retraining phase is to minimize the loss:
\begin{equation}
    \mathcal{L}' = \mathcal{L}(\Theta, X, Y) + \gamma\sum_l\sum_i^{c_lc_{l+1}}||\alpha_i^{(l)}||_1,
\end{equation}
where the first term is the original loss of the model (i.e., cross entropy loss), the second term is the sum of the coefficients magnitude and $\gamma$ is the strength of the sparsity regularization.

If we visualize these two parameter sets as separate layers, it conceptually increases the depth of the model and makes it harder to converge. Thus, in the training process, we generate the reconstructed parameter $\hat{\theta}$ from the basis and coefficients and compute the gradients as the original convolutional layer. The chain rule can then be applied to derive the gradients of the basis and the coefficients from the original convolutional layer's gradients. Specifically,

\begin{equation}
\frac{\partial \mathcal{L}'}{\partial \mathbf{B}} =  (\frac{\partial\mathcal{L}'}{\partial\hat{\theta}})^T\mathbf{A},\quad\quad\frac{\partial \mathcal{L}'}{\partial \mathbf{A}} =  \frac{\partial\mathcal{L}'}{\partial\hat{\theta}}\mathbf{B}^T + \gamma,
\end{equation}
where the $\hat{\mathbf{\theta}} = \mathbf{AB}^T$ and $\mathbf{A}\in\mathbb{R}^{c_lc_{l+1}\times d}$ is the coefficient matrix. Again, we omit the superscript $l$ for readability. 

The gradient of coefficient matrix consists of two terms. The first term pushes the coefficient towards the direction that decreases the error, while the second term coerces the reconstructed kernels to be close to the basis kernels. If we jointly train the basis and coefficients, the coefficients will be updated based on the old basis and vice versa. Jointly training both makes it very difficult for the model to converge, producing further accuracy drop. We avoid this issue by conducting retraining in an alternating fashion, i.e., freezing the coefficients and train the basis for several epochs and then freezing the basis and train coefficients.

The decomposed manner can also benefit the sparsity regularization. Since $l1$-regularizer is non-smooth, it equates to adding a constant to the gradient, which dominates the gradient in the later stage of training. This causes the training process to be very unstable or unable to converge. Regularization on the decomposed filter state avoids this issue. Examining the gradient from the original weight's perspective, the regularization constant is essentially scaled as a consequence of being applied only to the coefficients. This scaling factor decreases the constant proportional to the diminishing gradients. The constant is still within the same magnitude of the gradient of the loss term, thus stabilizing the process of converging to a sparse parameter set.
 
\subsection{Model Shrinking}
\label{sec:prune}
Retraining the filter-decomposed model with sparsity regularization results in predominantly near-zero coefficients. As shown in \cite{han2015deep}, we can select a threshold based on the standard deviation of each coefficient matrix and prune all weight values with a magnitude lower than the threshold. Only a few epochs of coefficient fine-tuning is required to recover accuracy lost by pruning. A combination of high sparsity level and low accuracy loss can be achieved without any additional iterations.
\begin{figure}[b]
    \centering
    \includegraphics[width=\linewidth]{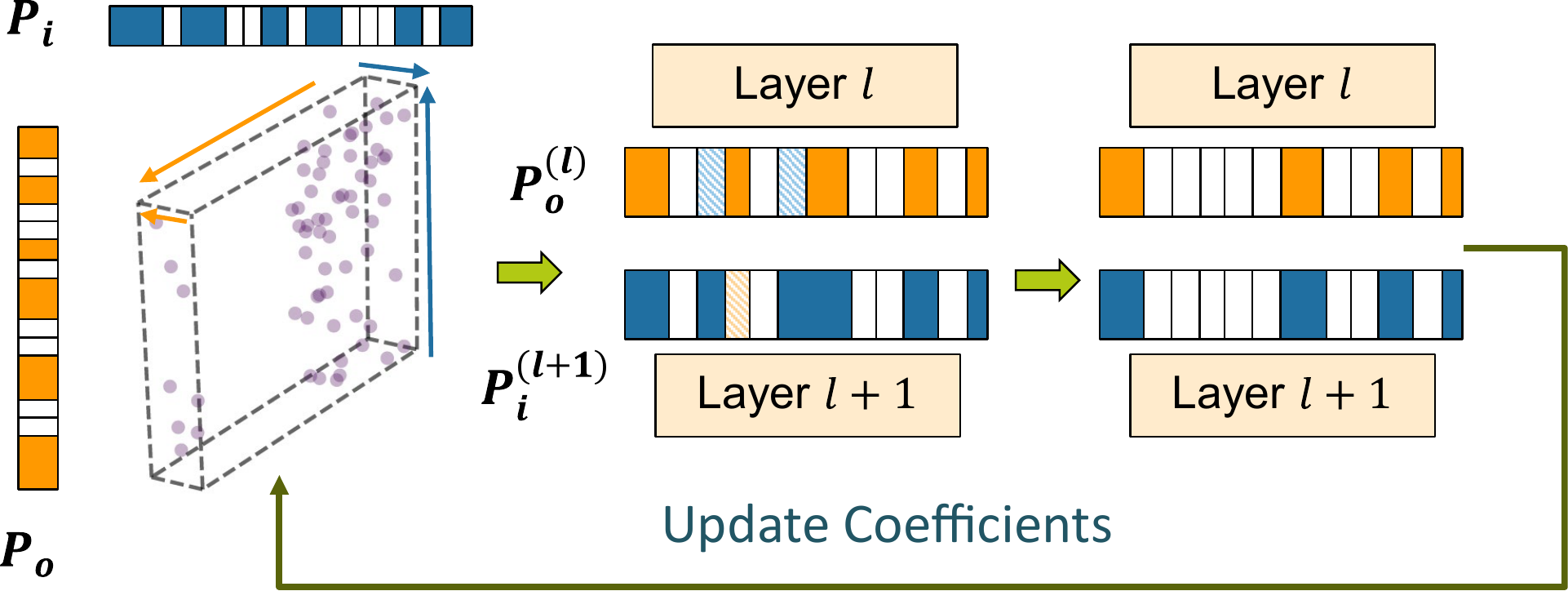}
    \caption{Model shrinking procedure. The blank items in $P_o^{(l)}$ and $P_i^{(l+1)}$ represents the redundant channels, while the shaded items denote the difference of the two redundant sets. }
    \label{fig:shirnkage}
\end{figure}

The sparse coefficients expose redundancies in CNN structures that can be utilized to shrink the model. Model shrinkage begins with reshaping the coefficient matrix $\theta^{(l)}$ into the shape $c_l\times c_{l+1} \times k'$. By selecting the first dimension (i.e., the input channels) and summing the number of the non-zero elements of the remaining two dimensions, we can obtain a vector $p_{i}^{(l)}$ with $c_l$ elements. Zeros in $p_{i}^{(l)}$ indicate that corresponding input channels are redundant since no output channels are connected. Indices of these channels can be represented by the set $P_{i}^{(l)}$. Selecting the second dimension (i.e., the output channels) and conducting the same procedure, we can get $p_{o}^{(l)}$ and $P_{o}^{(l)}$, which indicate redundant output channels. The redundancies in basis kernels can also be derived with the same procedure.

Note that it is possible for redundancies of a layer's input and output channels to not match. We can exploit this feature by considering the connections between input channels and redundant output channels of the same layer. If some input channels only have connections to redundant output channels, these inputs consequentially become redundant. Thus, we iteratively update the redundancy sets by applying the following steps. First, we take the union of the current layer's output channels with the next layer's input channels, i.e., $P_o^{(l)} \longleftarrow  P_i^{(l+1)}  \longleftarrow P_i^{(l)}\cup P_o^{(l+1)}$. Then, we update $\theta^{(l)}$ by setting all corresponding coefficients in $P_o$ and $P_i$ to zero and deriving new redundancy vectors and sets. This procedure, depicted in Fig. \ref{fig:shirnkage}, is repeated until no modification is made in an iteration.

A potential problem with the iterative $\theta^{(l)}$ update procedure is when it is applied to CNN architectures with skip connections, such as ResNet \cite{he2016deep}. Specifically, dimensions of pruned output feature maps might be inconsistent with corresponding skip connections. The solution to this issue is straightforward. If the shortcut path has a dimension-matching operation (i.e., 1$\times$1 convolution), we update the output channel of the 1$\times$1 convolution and the current layer by taking the intersection of their redundancy sets. If the shortcut path has no such operation, we will need to update the redundancy sets of the start and the end of the skip connection before updating the coefficients.

\begin{figure}[hb]
    \centering
    \subfigure[With dimension matching component.]{
    \includegraphics[width=\linewidth]{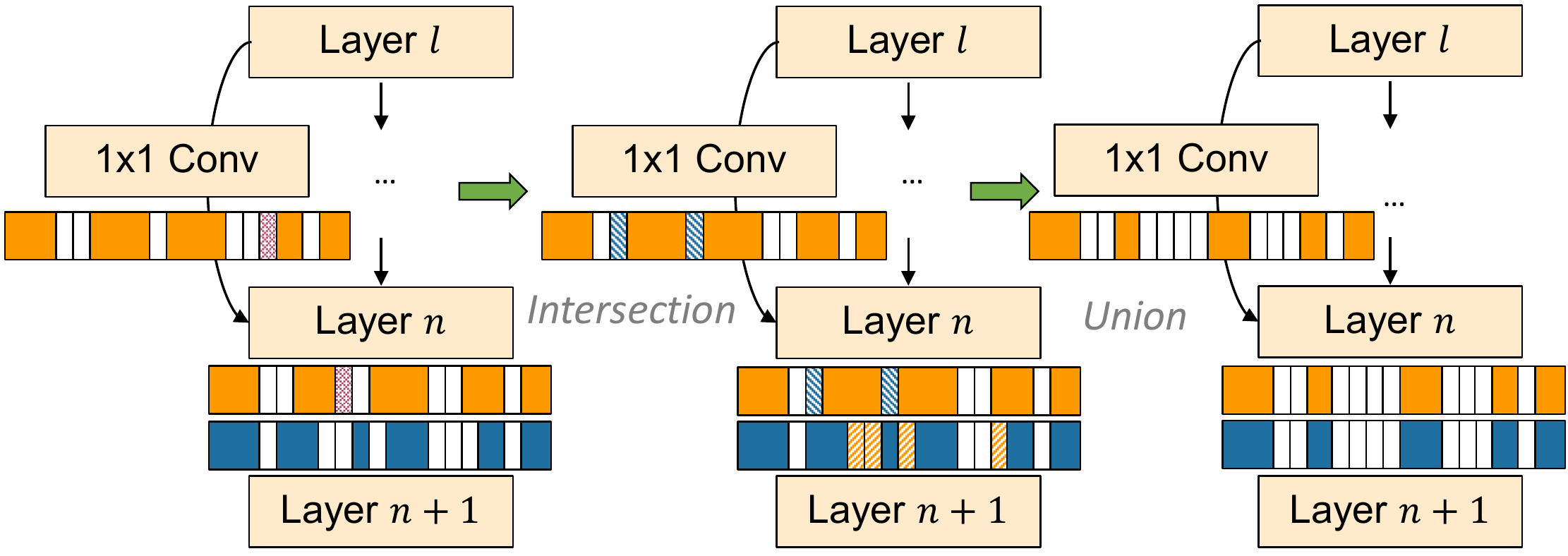}
    }
    \subfigure[Without dimension matching component.]{
    \includegraphics[width=\linewidth]{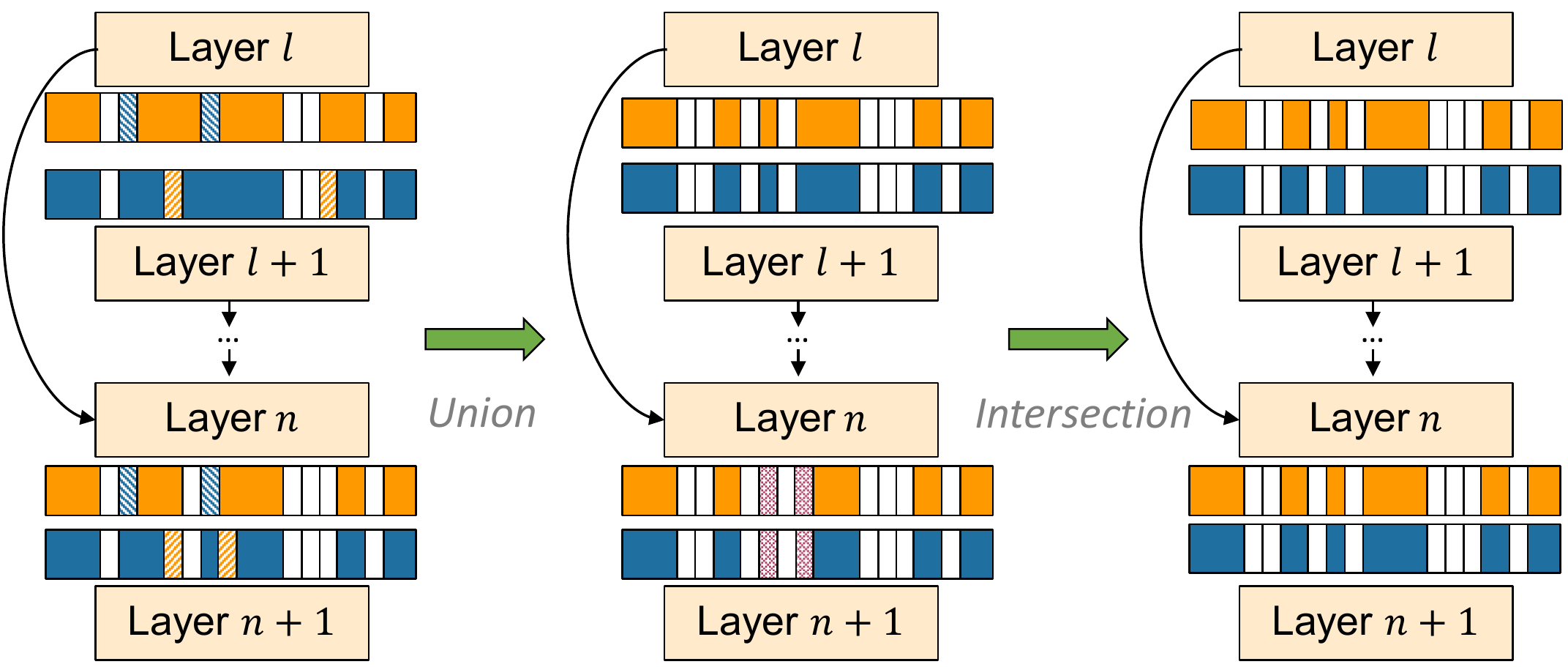}
    }
    \caption{Shrinking a model containing skip connections. The shaded items represent the difference of the redundant sets in each step. The corresponding items will be eliminated (added) in the intersection (union) step.}
    \label{fig:shrink_res}
\end{figure}

\subsection{Hardware Benefit}
\label{sec:hd_ben}

The decisive advantage of PENNI over previous CNN pruning, compression, or filter decomposition methods is its potential for synergistic reduction of memory and computational footprints. PENNI directly leverages filter decomposition by enabling a partition of the convolution step into two distinct stages.

The first stage involves channel-by-channel convolutions with each of the $d$ two-dimensional basis kernels, producing $c_ld$ intermediate feature maps; this stage is analogous to to depthwise separable convolution \cite{chollet2017xception} with $d$ branches. Each branch duplicates one of the basis kernels across the $c_l$ input channels. Applying such a technique greatly reduces the number of multiply-and-accumulates (MACs) in the convolution step, which is the bottleneck in convolutional layers.

The second stage is a weighted sum to produce the convolutional layer's output feature map. Specifically, $c_ld$ intermediate feature maps are multiplied element-wise with the coefficient matrix and then accumulated at the output. As described in Section \ref{sec:prune}, the coefficient matrices are incredibly sparse; therefore, we reduce the model's memory footprint and prevent redundant zero-multiply computations by representing the coefficients through a sparse matrix format. Although this stage introduces additional computations that offset the reduction in MACs from the first stage, the overall number of computations is dramatically reduced, thus improving inference latency. 

Beyond the aforementioned straightforward benefits of the proposed two-stage convolutional layer scheme, PENNI also offers a unique attribute that can be leveraged for current and future hardware accelerator designs. The deterministic convolutional kernel structure means that the number of basis kernels can be altered to fit nicely with the number of processing elements (PEs) in accelerators such as DaDianNao \cite{chen2014dadiannao} without forcing the model to conform to the hardware (e.g. reducing layer width). Meanwhile, the weighted sum stage can be computed in a streaming manner, much favored by single-instruction, multiple-data (SIMD) processors. Also, because data access patterns of convolutional layers conventionally require hardware-specific data-reuse algorithms to minimize costly cache evictions, removing interactions of the input channels at the convolution step via depthwise separation alleviates hardware complexity. Lastly, partitioning the convolution step to two stages opens the avenue for further accelerator-based throughput optimizations such as pipelining. 
\begin{table*}[t]
\caption{Compression Result on CIFAR10. \textit{`Ours-D'} denotes the result with only the decomposition and retraining (i.e., phase A and phase B in Figure \ref{fig:framework}), while \textit{`Ours-P'} incorporates the pruning and model shrinkage based on the \textit{`Ours-D'} model. `-' denotes unavailable data from the original paper.}
\label{tab:cifar10}
\vskip 0.15in
\begin{center}
\begin{small}
\begin{tabular}{lcccccccr}
\toprule
Arch & Method & Base Acc. & Pruned Acc. &$\Delta_{Acc}$ & Param. & $R_{Param.}$& FLOPs &$R_{FLOPs}$ \\
\midrule
VGG16 & Baseline & 93.49\%& - & -& 14.71M & - & 314.26M & - \\
 & PFEC & 93.25\%& 93.40\%& -0.15\%& 5.4M & 64\% & 206M & 34.2\% \\
 & Slimming & 93.62\%& \textbf{93.56\%}& \textbf{-0.06\%}& 1.77M & 87.97\% & 127M & 43.50\% \\
  & AOFP & 93.38\%& 93.28\%& -0.10\%&- & - & 77M & 75.27\% \\
 & Ours-D & 93.49\%& 93.14\%& -0.35\%&183.4M & 44.44\% & 183.4M &41.65\% \\      
 & Ours-P & 93.49\%& 93.12\%& -0.37\%&\textbf{0.135M}& \textbf{98.33\%}&\textbf{21.19M} & \textbf{93.26\%}  \\       

\midrule
ResNet18 & Baseline & 93.77\%& - &-& 11.16M &-& 555.43M&- \\

 & Ours-D & 93.77\% & 93.89\%& +0.12\%& 6.28M& 56.27\%& 332.34M & 40.17\%  \\   
 & Ours-P & 93.77\%& \textbf{94.01}\%& \textbf{+0.24\%}& \textbf{0.341M}&\textbf{96.94\%}& \textbf{44.98M}&\textbf{91.90}\%  \\   

\midrule
ResNet56 & Baseline & 93.57\%& - &-& 0.848M&- & 125.49M&-\\
 & PFEC & 93.04\%& 93.06\%& +0.02\%& 0.73M & 13.7\%& 90.9M&27.6\% \\
 & SFP & 93.59\%&  93.35\%& -0.24\%&-&- & 59.4M&52.67\%\\
 & C-SGD & 93.39\%&  93.44\%& +0.05\%&-&- &-& 60.85\%\\
 & FPGM & 93.59\%&  93.49\%& -0.10\%& -&- & 59.4M&52.67\%\\
 & Group-HS & 93.14\%& 93.45\%& +0.31\%& -&-&-& 68.43\%\\
 & Ours-D &93.57\%& \textbf{94.00\%}& \textbf{+0.43\%}& 0.471M&44.46\%&92.80M& 26.15\%  \\   
 & Ours-P & 93.57\%& 93.38\%& -0.19\%& \textbf{39.37K}& \textbf{95.36\%}&  \textbf{28.98M}& \textbf{79.40\%}  \\   

\bottomrule
\end{tabular}
\end{small}
\end{center}
\vskip -0.1in
\end{table*}
\section{Experiments}
\label{sec:exp}
In this section, we demonstrate the effectiveness of the proposed framework. Experiments were held on CIFAR10 \cite{krizhevsky2009learning} and ImageNet \cite{deng2009imagenet} datasets. Experiment settings are detailed before comparing compression results between PENNI and both state-of-the-art channel pruning and weight pruning methods. Finally, we conduct an ablation study to show the contribution of each component in the framework.

\subsection{Experiment Settings}

\paragraph{CIFAR10} On CIFAR-10, we chose VGG16 \cite{simonyan2014very}, ResNet18 and ResNet56 \cite{he2016deep} for experimentation. VGG16 is a small version tailored for CIFAR10. We use ResNet56 to test the performance on compact models. Model training involved the following data preprocessing steps: random flipping, random cropping with 4 pixels padding, and normalization. The VGG16 and ResNet18 models were first pretrained for 100 epochs with 0.1 initial learning rate; then, the learning rate was multiplied by 0.1 at 50\% and 75\% epochs, while ResNet56 was pretrained for 250 epochs with the same learning rate scheduling. All pretraining, retraining and fine-tuning procedures implemented Stochastic Gradient Descent (SGD) as the optimizer with $10^{-4}$ weight decay, 0.9 momentum, and batch size set to 128. We selected $d=5$ for the decomposition stage and retrained for 100 epochs with 0.01 initial learning rate and the same scheduling. Regularization strength was set to $\gamma=10^{-4}$. The interval between training basis and coefficients was set to 5 epochs. The final fine-tuning procedure took 30 epochs with 0.01 initial learning rate and the same scheduling scheme. 

\paragraph{ImageNet} On ImageNet, we used AlexNet \cite{krizhevsky2012imagenet} and ResNet50 for the experiment, incorporating the pretrained models provided by PyTorch \cite{pytorchmodel}. Since AlexNet has different kernel sizes across layers, we selected $d=64$ and $14$ for the first two convolutional layers, and $d=5$ for the rest 3$\times$3 convolutional layers. For ResNet50, we use 4 sets of parameter settings, with $d=5, 6$ and regularization strength set to $5e-5$ and $1e-4$. The retraining procedure lasted 50 epochs with the same hyper-parameters as CIFAR10 but set batch size to 256 and cosine annealing.  For AlexNet, we warmed up with a learning rate of $0.0001$ for five epochs; then, the learning rate was set to $0.001$ for the remaining 45 epochs. The fine-tune procedure took 30 epochs with learning rate set to 0.01 for ResNet50 and 0.0001 for AlexNet.

\subsection{CIFAR10 Results}
\label{sec:cifar10_exp}

We selected channel pruning methods PFEC \cite{li2016pruning}, Slimming \cite{liu2017learning}, SFP \cite{he2018soft}, AOFP \cite{ding2019approximated}, C-SGD \cite{ding2019centripetal}, FPGM \cite{he2019filter} and Group-HoyerSquare \cite{yang2019deephoyer}  for comparison. For the works providing parameter trade-offs, we use results with similar accuracy drop. The results are shown in Table \ref{tab:cifar10}. `Ours-D' denotes the compression result with only decomposition and retraining phases, while `Ours-P' incorporates all four phases. We only consider the parameters of the convolutional and linear layers, and the FLOP count is taken by calculating the number of Multiply-Accumulation (MAC) operations. Based on the computation flow described in Section \ref{sec:hd_ben}, we consider that the sparsity of coefficient matrix can be converted to reduction in FLOPs. Thus, we ignore the zeros in the coefficient matrices when counting FLOPs. On VGG16, we outperformed channel pruning methods by achieving a reduction over 98\% on parameters and 93.26\% on FLOPs. Although there is a slightly higher accuracy drop, it is only 0.15\% behind AOFP with 10\% extra reduction on FLOPs and 0.42\% behind Slimming with almost double reduction on FLOPs, which is acceptable. Since ResNet18 is originally designed for the ImageNet dataset, no previous work has provided result for comparison. We include it in this paper to show that PENNI is able to shrink over-parameterized models and may improve accuracy. On ResNet56, which is a compact model specially tailored for CIFAR10, we can still prune 94.52\% parameters and 76.9\% FLOPS with  0.2\% accuracy drop. Our method outperformed previous channel pruning methods by a nearly 20\% extra FLOPs reduction, and a 10\% extra reduction over the group regularization method.

\subsection{ImageNet Results}
\begin{table}[h]
\caption{Compression Result of AlexNet on ImageNet.}
\label{tab:imagenet_alex}

\begin{center}
\begin{small}
\begin{tabular}{lcccr}
\toprule
 Method & Top-1 & Top-5 & FLOPs&$R_{FLOPs}$ \\
\midrule
 Baseline & 56.51\%& 79.07\% & 773M & - \\ 
 AOFP &  \textbf{56.17\%}&  \textbf{79.53\%} & 492M &41.33\%\\
 Ours-D &  55.41\%&  78.30\%& 573M &25.88\% \\   
 Ours-P &  55.57\%&  78.32\%& \textbf{232M} &\textbf{70.04\%} \\   
\bottomrule
\end{tabular}
\end{small}
\end{center}
\end{table}

\begin{table}[h]
\caption{Compression Result of ResNet50 on ImageNet. We categorize the results by accuracy.}
\label{tab:imagenet_res}

\begin{center}
\begin{small}
\begin{tabular}{lcccr}
\toprule
Method & Top-1 & Top-5 & FLOPs &$R_{FLOPs}$ \\
\midrule
Baseline & 76.13\%& \textbf{92.86\%} & 4.09G & - \\   
\textbf{Ours-D} &  \textbf{76.20\%}&  92.85\% &   \textbf{3.23G} & \textbf{21.10\%}   \\   

\hline
ThiNet-70 &  72.02\%& 90.67\% & 2.58G\tablefootnote{\label{fn:flops}Computed based on the reduction percentage reported by original paper.}&36.80\%\\  
\textbf{Ours-R1} &  \textbf{73.87\%}&  \textbf{91.79\%} &  \textbf{220M} &\textbf{94.73\%}   \\

\hline 
SFP &  74.61\%&  92.06\% & 2.38G$^{\ref{fn:flops}}$&41.80\%\\ 
C-SGD-50 &  74.54\%&  92.09\% & 1.81G$^{\ref{fn:flops}}$&55.76\%\\ 
\textbf{Ours-R2} &  \textbf{74.74\%}&   \textbf{92.27\%} &   \textbf{527M} & \textbf{87.12\%}   \\

\hline
C-SGD-60 &  74.93\%&  92.27\% & 2.20G$^{\ref{fn:flops}}$&46.24\%\\ 
FPGM-40\% & 74.83\%&  \textbf{92.32\%} &  1.90G$^{\ref{fn:flops}}$ &53.50\%\\ 
 \textbf{Ours-R3} &  \textbf{75.00\%}&  92.21\% &   \textbf{576M} &\textbf{85.92\%}   \\
 
\hline
AOFP-C1 & 75.63\%&  92.69\% & 2.58G&32.88\%\\ 
AOFP-C2 & 75.11\%&  92.28\% & 1.66G&56.73\%\\ 
C-SGD-70 &  75.27\%&  92.46\% & 2.59G$^{\ref{fn:flops}}$&36.75\%\\ 
FPGM-30\% & 75.59\%&  92.63\% &  2.36G$^{\ref{fn:flops}}$ &42.20\%\\ 
\textbf{Ours-R4} &   \textbf{75.66\%}&   \textbf{92.79\%} &   \textbf{768M} &\textbf{81.23\%} \\
\bottomrule
\end{tabular}
\end{small}
\end{center}

\vskip -0.1in
\end{table}

On ImageNet, we chose Slimming, ThiNet \cite{luo2017thinet}, SFP, AOFP, C-SGD and FPGM for comparison. The result of AlexNet compression is shown in Table \ref{tab:imagenet_alex}. We can prune 70.04\% FLOPs with 1\% loss on top-1 accuracy.  For ResNet50, we observe the 1$\times$1 convolutional layer of the bottleneck block as the coefficient matrix with 1-D basis and apply regularization to it. Table \ref{tab:imagenet_res} shows the result on ResNet50 compression. We use multiple parameter settings to justify the trade-off between accuracy and compression rate. `Ours-D' only involves decomposition and retraining step with $d=5$, while `R1' and `R2' incorprate pruning and shrinking phases with regularization strength set to $1e-4$ and $5-e5$. `R3' and `R4' has the same parameter apart from setting  $d=6$ in the decomposition phase.  The results show that the decomposition step can reduce more than 20\% of the FLOPs with no accuracy drop. With the pruning and shrinking procedures, 94.73\% of the FLOPS can be pruned with 2.4\% top-1 accuracy loss. When we relax the regularization, we can still prune 81.23\% of the FLOPs with only 0.5\% accuracy loss. The FLOPs reduction is nearly two times the reduction of previous channel pruning methods. A even larger compression rate can be achieved by combining the 1$\times$1 convolutional layer with the coefficient matrices.

\subsection{Inference Acceleration}

\begin{table}[ht]
\caption{Measured inference performance of VGG16-CIFAR10 on different devices.}
\label{tab:inf}
\vskip 0.15in
\begin{center}
\begin{small}
\begin{tabular}{lcccr}
\toprule
Device & Variation & Latency(ms) & Memory(MB) \\
\midrule
CPU & Baseline & 12.9& 137\\
 & PENNI & 5.96& 77.6\\
\midrule
  GPU& Baseline & 10.8& 487\\
 & PENNI & 7.26& 424\\
\bottomrule
\end{tabular}
\end{small}
\end{center}
\vskip -0.1in
\end{table}

\paragraph{Hardware Settings} We used Intel Xeon Gold 6136 to test the inference performance for CPU platform and NVIDIA Titan X for the GPU platform. For software, we used PyTorch 1.4 \cite{paszke2019pytorch} to implement the inference test. Batch size was set to 128 (1) for inference testing on the GPU (CPU). GPU inference batch size is higher than CPU to increase utilization and emphasize the latency impact of our method on the highly parallel platform. We indicate these settings as latency and peak memory consumption values vary across platforms or library versions.

Table \ref{tab:inf} displays inference latencies and memory consumption recorded for the baseline and PENNI framework. 
As mentioned in \ref{sec:hd_ben}, one of PENNI's defining strengths is its impact on computational and memory footprints. Results shown in Table \ref{tab:inf} reveal a 1.5$\times$ (2.2$\times$) reduction in measured inference latency on the GPU (CPU). Peak memory consumption also benefited from a 1.1$\times$ (1.8$\times$) reduction. It is important to note that these metrics were taken without applying the convolution computation reorganization described in \ref{sec:hd_ben}; this is done intentionally to reveal the effectiveness of our model shrinkage with zero changes to the hardware and inference-time computation. The reduction in memory is a straightforward consequence of the decomposition and shrinking stages of the PENNI framework. Although our method is successful at dramatically decreasing model size in memory, intermediate feature maps seems to dominate on-device memory consumption, especially with a batch size of 128 on the GPU.

\subsection{Subspace Dimension}
\label{sec:dim}
\begin{figure}[htb]
    \centering
    \includegraphics[width=0.95\linewidth]{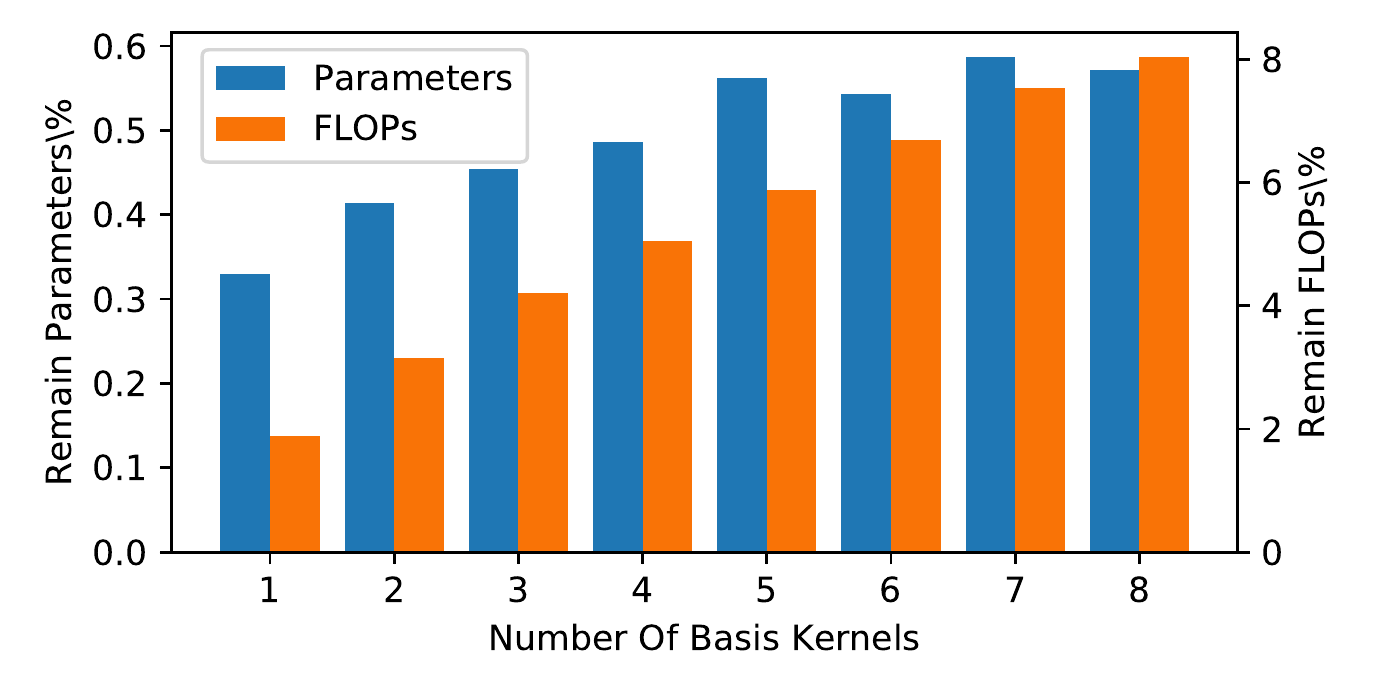}
    \includegraphics[width=0.95\linewidth]{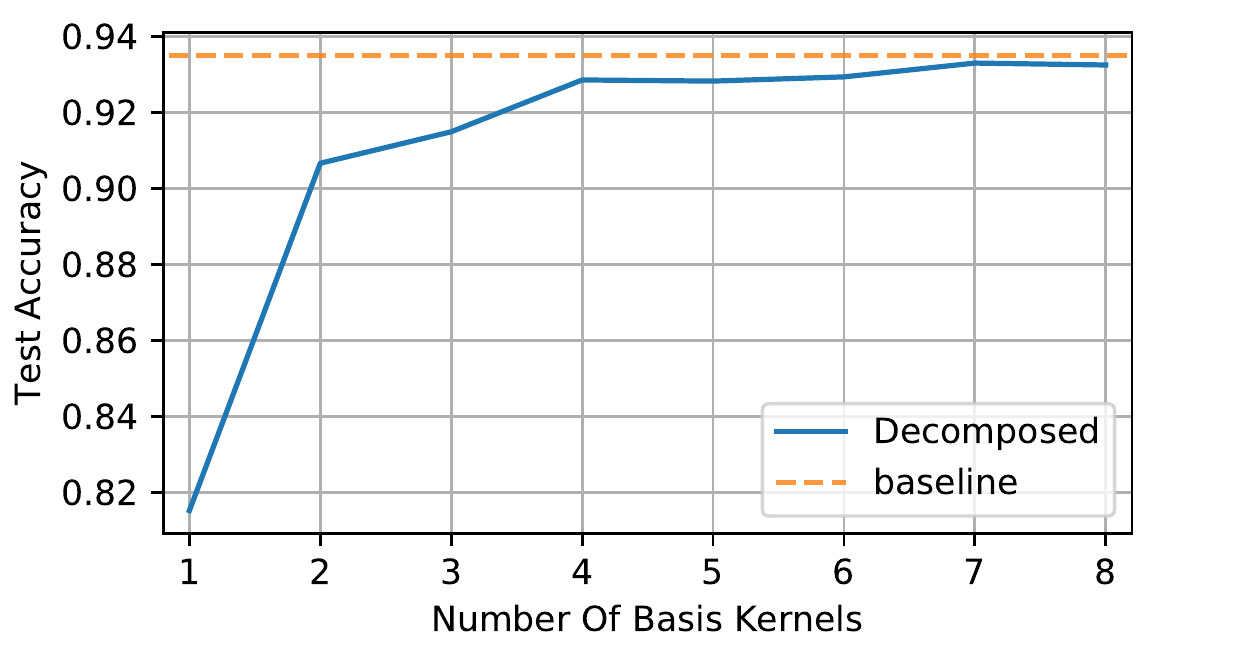}
    \caption{Test accuracy, parameters and computation reduction with different number of basis kernels $d$. }
    \label{fig:dim}
\end{figure}
To justify the selection of the parameter $d$, we conduct an experiment with different decompose dimensions. We used the same VGG16 baseline model and hyper-parameters as \ref{sec:cifar10_exp}. The result is shown in Fig.\ref{fig:dim} indicates that the remaining FLOPs scales linearly with the number of basis kernels. This is expected since the number of convolutional operations is determined by $d$. The parameters scale linearly before 6-D basis and have minor difference with the increasing dimension. This is because even though more basis vector requires more coefficients, it also adds flexibility and thus leads to sparser coefficients. The test accuracy reveals the same trend, with $d\geq 4$, minor improvement on the accuracy can be brought by increasing $d$. Thus, we select $d=5$ for the balance between parameter and FLOPs reduction and accuracy drops.

\subsection{Model Shrinking}

\begin{figure}[htb]
    \centering
    \subfigure[VGG16-CIFAR10]{
    \includegraphics[width=0.8\linewidth]{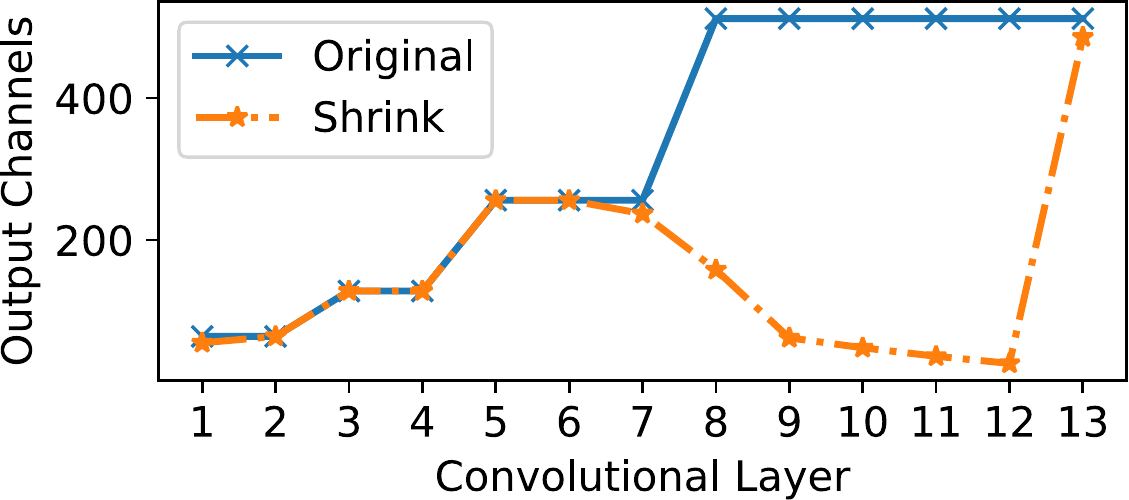}
    }
    \subfigure[ResNet56]{
    \includegraphics[width=0.8\linewidth]{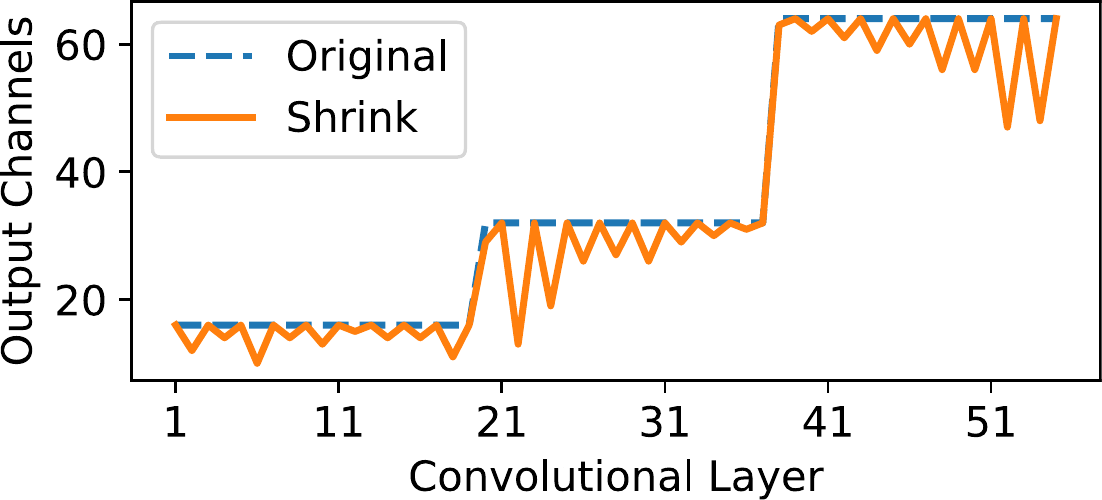}
    }
    \caption{Layer width after the model shrinking. }
    \label{fig:mdl_shr}
\end{figure}
We show the effectiveness of model shrinking by comparing layer widths. As shown in Figure \ref{fig:mdl_shr}, on VGG16, the model shrinking procedure effectively removes redundant channels in the second half of all layers. Meanwhile, on ResNet56, the shrinking is limited by the dimension matching requirement of the skip connections. The oscillation pattern of the layer width indicates that redundancies of the inner-block layer can be effectively exploited. These results show that PENNI can benefit unmodified inference software and hardware by exploiting structural redundancies.

\section{Conclusion}
\label{sec:con}

This work proposes the PENNI framework for hardware-friendly CNN model compression. Our method improves inference latency with no changes to inference algorithms and hardware via model shrinking, thus translating model sparsity to speedup. A low rank assumption is used to decompose CNN filters into basis kernels and prune the resulting coefficient matrices, which results in structured sparsity. A novel alternating fine-tuning method is used to further increase sparsity and improve model performance. Unique characteristics generated by the decomposition step may be leveraged for hardware efficiency via convolution computation reorganization, directly benefiting modern DNN platforms. 

\section*{Acknowledgment}

This work was supported in part by NSF-1937435, NSF-1822085,  NSF-1725456, ARO W911NF-19-2-0107, and NSF IUCRC for ASIC memberships from Cadence etc.

\bibliography{ref}
\bibliographystyle{icml2020}
\clearpage
\onecolumn
\section*{Appendix}
\section*{Visualization of coefficient matrix}

\begin{figure}[htb]
    \centering
    \subfigure[Decomposed]{
    \includegraphics[width=0.9\linewidth]{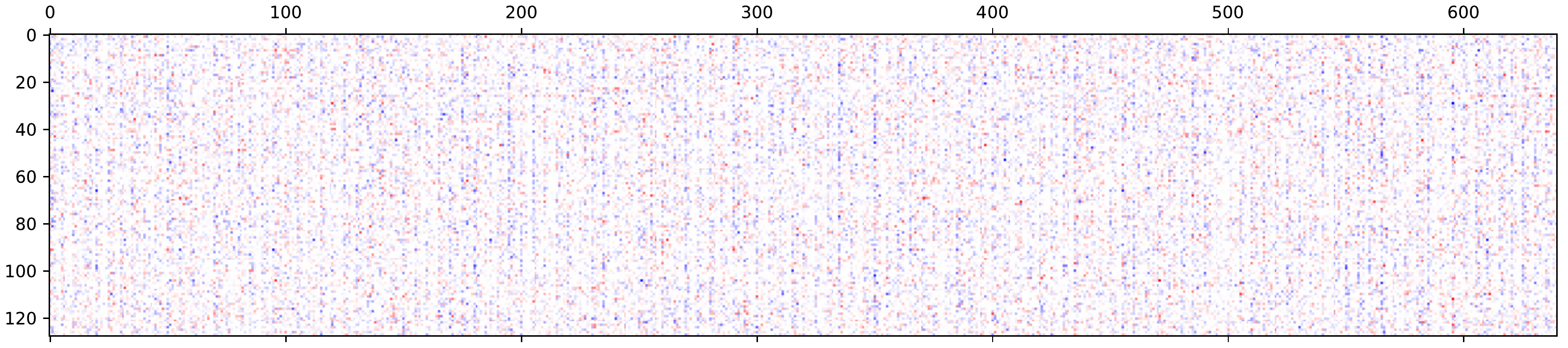}
    }
    \subfigure[Pruned]{
    \includegraphics[width=0.9\linewidth]{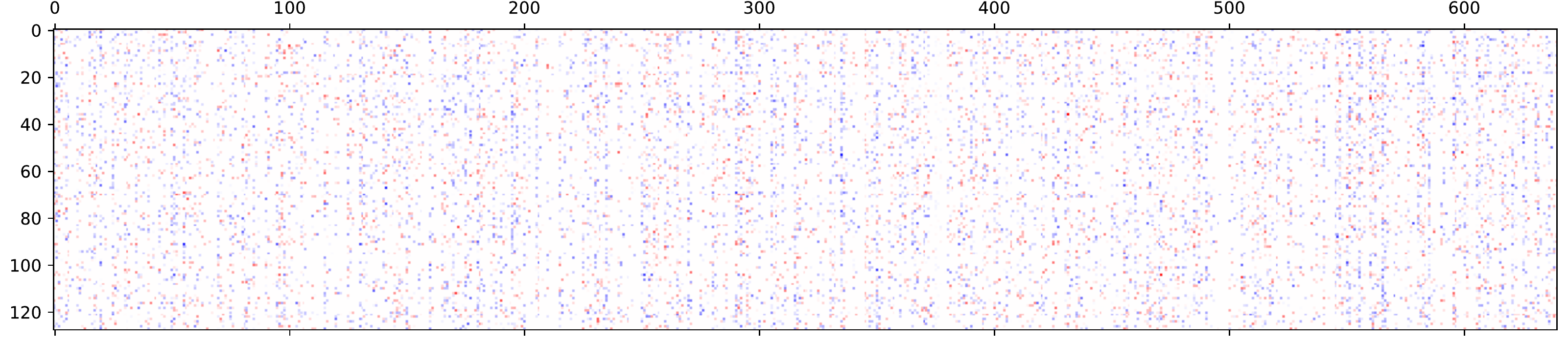}
    }
    \caption{The visualization of the coefficient matrix. This is the third convolution layer of VGG16-CIFAR10 model with 128 input channels and 128 convolutional filters. The y-axis represents the output channels while the x-axis is the basis of each input channel.}
    \label{fig:coef}
\end{figure}

The coefficient matrix is shown in Fig.\ref{fig:coef}. By conducting retraining, pruning and fine-tuning process, both weight and structured redundancies are explored.

\begin{figure}[htb]
    \centering
    \subfigure[Baseline]{
    \includegraphics[width=0.98\linewidth]{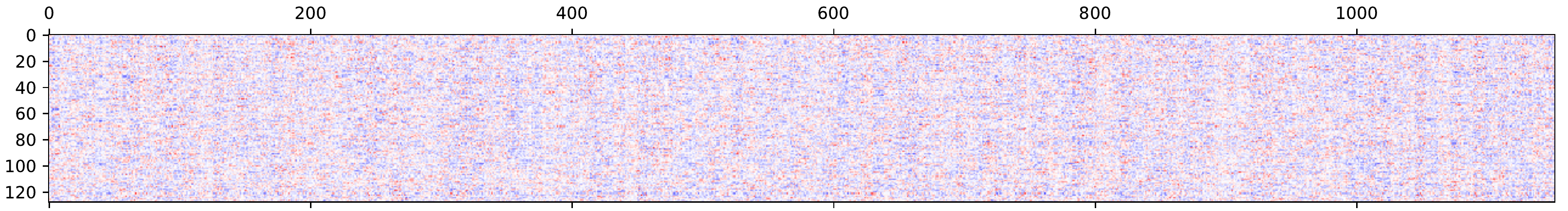}
    }
    \subfigure[Decomposed]{
    \includegraphics[width=0.98\linewidth]{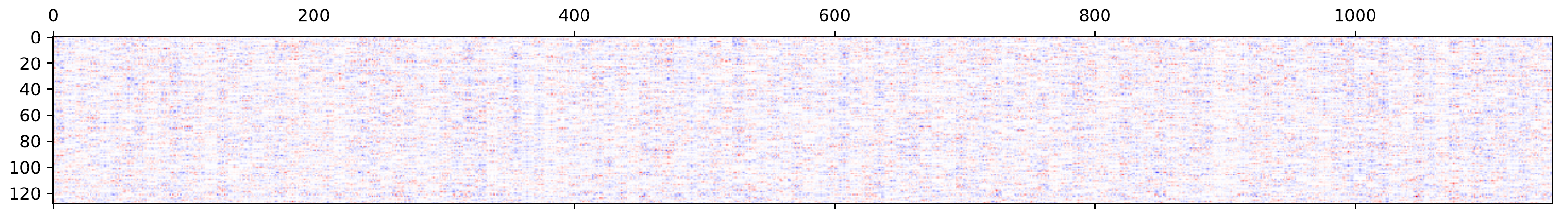}
    }
    \subfigure[Pruned]{
    \includegraphics[width=0.98\linewidth]{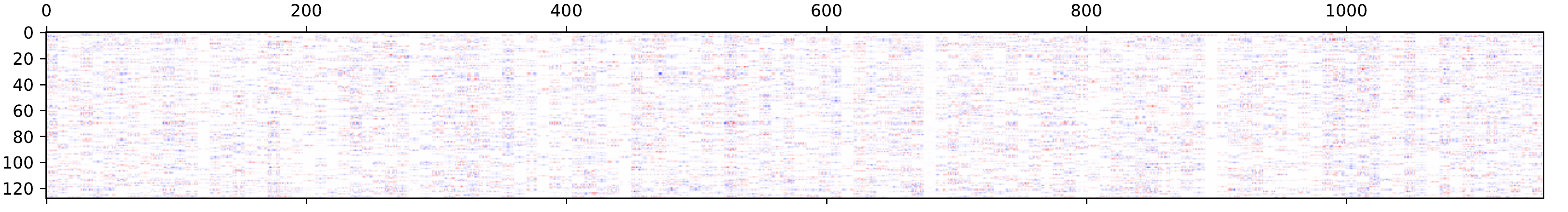}
    }
    \caption{The visualization of the reconstructed weight matrix comparing with the baseline model. This is the third convolution layer of VGG16-CIFAR10 model with 128 input channels and 128 convolutional filters. The y-axis represents the output channels.}
    \label{fig:recon}
\end{figure}

To compare with the baseline model, we reconstruct the weight matrix by multiplying the coefficient matrix with the basis. The reconstructed weight matrix is shown in Fig.\ref{fig:recon}. Although the sparsity is also shown in the reconstructed weight, more computation can be saved by using the decomposed convolution.

\section*{Pruned Model Structure}
In this section, we list the detailed structure of the pruned model, including the number of basis kernels, the layer width before/after the shrinking process and the sparsity level of the coefficient matrix. We only consider the convolutional layers.

\subsection*{VGG16 on CIFAR10}
We list the detailed structure of the VGG16 before and after the shrinking process on Table.\ref{tab:vgg}. We only consider all the convolutional layers. The first index of the name represents the downsampling stage. The width represents the number of the filters (i.e., the number of output channels).

\begin{table}[H]
\caption{Detailed Structure of VGG16-CIFAR10}
\vspace{1em}
\label{tab:vgg}
\begin{tabular}{|c|c|c|c|c|c|c|c|c|c|}
\hline
\multirow{2}{*}{Name} & \multirow{2}{*}{Size} & \multicolumn{4}{c|}{Before shrinking}                                                                       & \multicolumn{4}{c|}{After shrinking}                                                                       \\ \cline{3-10} 
                      &                       & Width & Basis & \begin{tabular}[c]{@{}c@{}}Coefficients\\ (Non-zero/Total)\end{tabular} & Sparsity/\% & Width & basis & \begin{tabular}[c]{@{}c@{}}Coefficients\\ (Non-zero/Total)\end{tabular} & Sparsity/\% \\ \hline
conv1\_1              & 3                     & 64    & 5           & 196/960                                                                 & 79.58       & 55    & 5           & 196/825                                                                 & 76.24      \\ \hline
conv1\_2              & 3                     & 64    & 5           & 2313/20480                                                              & 88.71       & 64    & 5           & 2238/17600                                                              & 87.28      \\ \hline
conv2\_1              & 3                     & 128   & 5           & 5862/40960                                                              & 85.69       & 128   & 5           & 5862/40960                                                              & 85.68      \\ \hline
conv2\_2              & 3                     & 128   & 5           & 12052/81920                                                             & 85.29       & 128   & 5           & 12052/81920                                                             & 85.28      \\ \hline
conv3\_1              & 3                     & 256   & 5           & 23169/163840                                                            & 85.86       & 256   & 5           & 23169/163840                                                            & 85.85      \\ \hline
conv3\_2              & 3                     & 256   & 5           & 36870/327680                                                            & 88.75       & 256   & 5           & 36870/327680                                                            & 88.74      \\ \hline
conv3\_3              & 3                     & 256   & 5           & 27719/327680                                                            & 91.54       & 237   & 5           & 27716/303360                                                            & 90.86      \\ \hline
conv4\_1              & 3                     & 512   & 5           & 15688/65536                                                             & 97.61       & 158   & 5           & 15665/187230                                                            & 91.63      \\ \hline
conv4\_2              & 3                     & 512   & 5           & 4534/1310720                                                            & 99.65       & 62    & 5           & 4530/48980                                                              & 90.75      \\ \hline
conv4\_3              & 3                     & 512   & 5           & 2668/1310720                                                            & 99.79       & 48    & 5           & 2666/14880                                                              & 82.08      \\ \hline
conv5\_1              & 3                     & 512   & 5           & 1318/1310720                                                            & 99.89       & 36    & 5           & 1315/8640                                                               & 84.78      \\ \hline
conv5\_2              & 3                     & 512   & 5           & 874/1310720                                                             & 99.93       & 26    & 5           & 874/4680                                                                & 81.32      \\ \hline
conv5\_3              & 3                     & 512   & 5           & 2554/1310720                                                            & 99.80       & 486   & 5           & 2554/63180                                                              & 95.95      \\ \hline
Total                 &                       &       &             & 135817/8172480                                                          & 98.34       &       &             & 135707/1263775                                                          & 89.26      \\ \hline
\end{tabular}
\end{table}

\subsection*{ResNet50 on ImageNet}
We list the detailed structure of ResNet50-R1 after shrinking process in Table.\ref{tab:my-table}.  We only consider all the convolutional layers including the 1$\times$1 convolution. 'L' represents the downsampling stages while the 'B' stands for the residual blocks. This table reveals that, although for some layer the reduction of the layer width is not significant, the model can still benefit from pruning basis kernels.

\begin{longtable}[c]{|l|l|l|l|l|l|}
\caption{Detailed Structure of ResNet50 after the shrinking process.}
\label{tab:my-table}\\
\hline
Name & Size & width & \# of basis & \begin{tabular}[c]{@{}l@{}}coefficients\\ (non-zero/total)\end{tabular} & sparsity/\% \\ \hline
\endfirsthead
\multicolumn{6}{c}%
{{\bfseries Table \thetable\ continued from previous page}} \\
\hline
Nam & Size & width & \# of basis & \begin{tabular}[c]{@{}l@{}}coefficients\\ (non-zero/total)\end{tabular} & sparsity/\% \\ \hline
\endhead
conv1 & 7 & 59 & 5 & 128/885 & 85.53 \\ \hline
L1B1.conv1 & 1 & 22 & - & 500/1289 & 61.48 \\ \hline
L1B1.conv2 & 3 & 29 & 4 & 125/2552 & 95.1 \\ \hline
L1B1.conv3 & 1 & 256 & - & 2223/7424 & 70.06 \\ \hline
L1B1.downsample & 1 & 256 & - & 2589/15104 & 82.86 \\ \hline
L1B2.conv1 & 1 & 32 & - & 2636/8192 & 67.82 \\ \hline
L1B2.conv2 & 3 & 34 & 5 & 349/5440 & 93.58 \\ \hline
L1B2.con3 & 1 & 256 & - & 2301/8704 & 73.56 \\ \hline
L1B3.conv1 & 1 & 31 & - & 2978/7936 & 62.47 \\ \hline
L1B3.conv2 & 3 & 63 & 5 & 807/9756 & 91.74 \\ \hline
L1B3.conv3 & 1 & 256 & - & 3395/16128 & 78.95 \\ \hline
L2B1.conv1 & 1 & 88 & - & 5973/22528 & 73.49 \\ \hline
L2B1.conv2 & 3 & 115 & 3 & 581/30360 & 98.09 \\ \hline
L2B1.conv3 & 1 & 512 & - & 11939/58880 & 79.72 \\ \hline
L2B1.downsample & 1 & 512 & - & 23134/131072 & 82.35 \\ \hline
L2B2.conv1 & 1 & 20 & - & 3778/10240 & 63.11 \\ \hline
L2B2.conv2 & 3 & 76 & 4 & 334/6080 & 94.51 \\ \hline
L2B2.conv3 & 1 & 512 & - & 8154/38912 & 79.05 \\ \hline
L2B3.conv1 & 1 & 66 & - & 9898/33792 & 70.71 \\ \hline
L2B3.conv2 & 3 & 98 & 4 & 497/25872 & 98.08 \\ \hline
L2B3.conv3 & 1 & 512 & - & 10702/50176 & 78.67 \\ \hline
L2B4.conv1 & 1 & 78 & - & 12508/39936 & 68.68 \\ \hline
L2B4.conv2 & 3 & 95 & 4 & 1006/29640 & 96.61 \\ \hline
L2B4.conv3 & 1 & 512 & - & 12680/48640 & 73.93 \\ \hline
L3B1.conv1 & 1 & 240 & - & 29283/122880 & 76.17 \\ \hline
L3B1.conv2 & 3 & 229 & 2 & 712/100920 & 99.35 \\ \hline
L3B1.conv3 & 1 & 1024 & - & 57763/234496 & 75.37 \\ \hline
L3B1.downsample & 1 & 1024 & - & 117775/524288 & 77.54 \\ \hline
L3B2.conv1 & 1 & 103 & - & 29283/122880 & 76.17 \\ \hline
L3B2.conv2 & 3 & 182 & 5 & 866/93730 & 99.08 \\ \hline
L3B2.conv3 & 1 & 1024 & - & 57763/234496 & 75.37 \\ \hline
L3B3.conv1 & 1 & 94 & - & 35956/96256 & 62.65 \\ \hline
L3B3.conv2 & 3 & 180 & 5 & 1222/84600 & 98.56 \\ \hline
L3B3.conv3 & 1 & 1024 & - & 53268/184320 & 71.1 \\ \hline
L3B4.conv1 & 1 & 135 & - & 49843/138240 & 63.94 \\ \hline
L3B4.conv2 & 3 & 193 & 5 & 925/130275 & 99.29 \\ \hline
L3B4.conv3 & 1 & 1024 & - & 58240/197632 & 70.53 \\ \hline
L3B5.conv1 & 1 & 155 & - & 56279/158720 & 64.54 \\ \hline
L3B5.conv2 & 3 & 197 & 4 & 1138/122140 & 99.07 \\ \hline
L3B5.conv3 & 1 & 1024 & - & 60387/201728 & 70.07 \\ \hline
L3B6.conv1 & 1 & 206 & - & 65496/210944 & 68.95 \\ \hline
L3B6.conv2 & 3 & 217 & 4 & 1156/178808 & 99.35 \\ \hline
L3B6.conv3 & 1 & 2024 & - & 65842/222208 & 70.37 \\ \hline
L4B1.conv1 & 1 & 495 & - & 148203/506880 & 70.76 \\ \hline
L4B1.conv2 & 3 & 484 & 1 & 546/239580 & 99.77 \\ \hline
L4B1.conv3 & 1 & 2048 & - & 279619/991232 & 91.79 \\ \hline
L4B1.downsample & 1 & 2048 & - & 580155/2097152 & 72.34 \\ \hline
L4B2.conv1 & 1 & 411 & - & 276731/841728 & 67.12 \\ \hline
L4B2.conv2 & 3 & 445 & 4 & 1334/731580 & 99.82 \\ \hline
L4B2.conv3 & 1 & 2048 & - & 275688/911360 & 69.75 \\ \hline
L4B3.conv1 & 1 & 501 & - & 307600/1026048 & 70.02 \\ \hline
L4B3.conv2 & 3 & 484 & 3 & 1076/727452 & 99.85 \\ \hline
L4B3.conv3 & 1 & 2048 & - & 245290/991232 & 75.25 \\ \hline
total &  &  &  & 2.98M/12.98M & 77.0 \\ \hline
\end{longtable}

\section*{Learning Curves}

We show the learning curve of both the retraining and fine-tuning phase of the ResNet-50 model on ImageNet dataset. The curve here belongs to the `R4' setting, which is $d=6$ and regularization strength $\lambda=5e-5$

\begin{figure}[htb]
    \centering
    \includegraphics[width=0.47\linewidth]{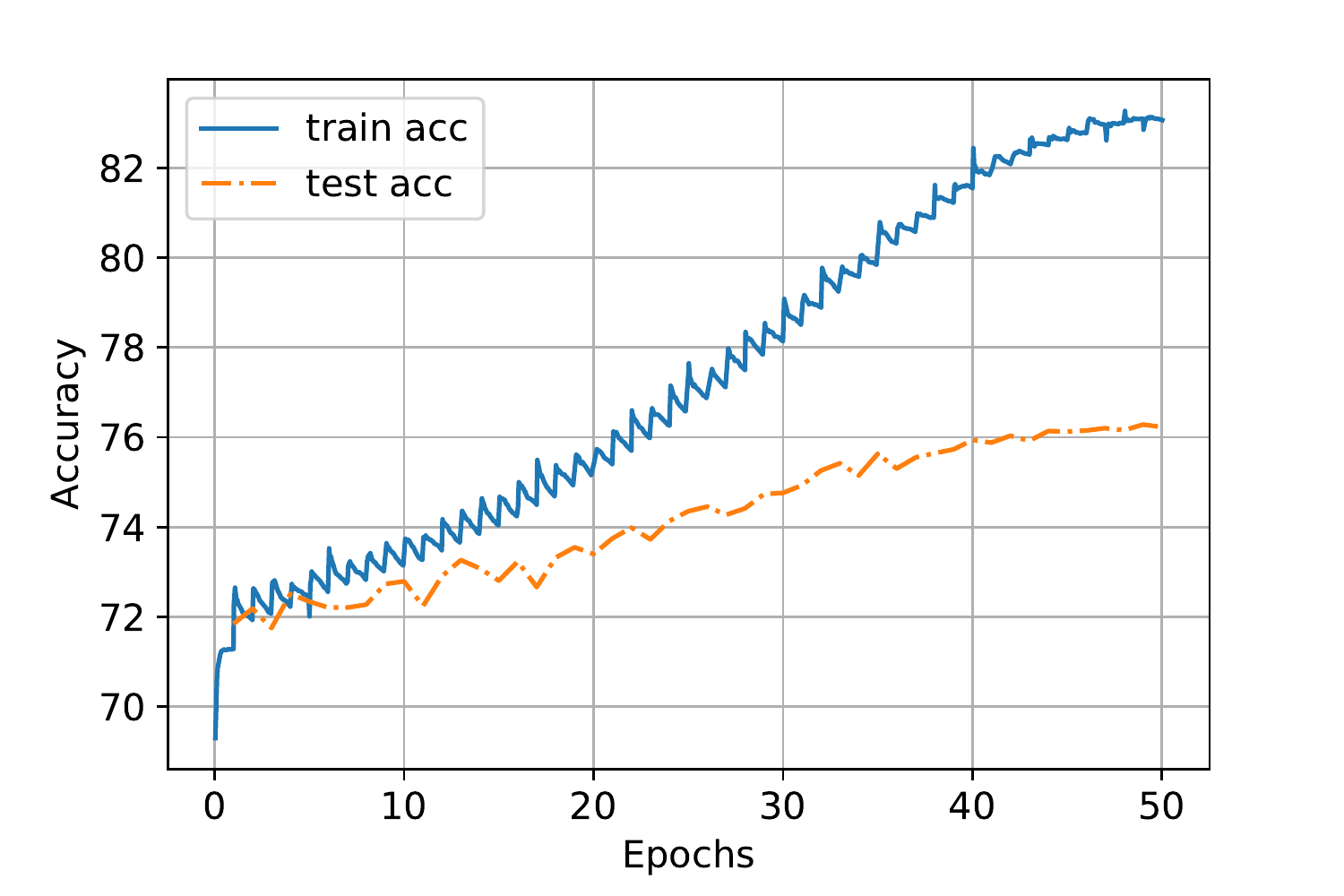}
    \includegraphics[width=0.47\linewidth]{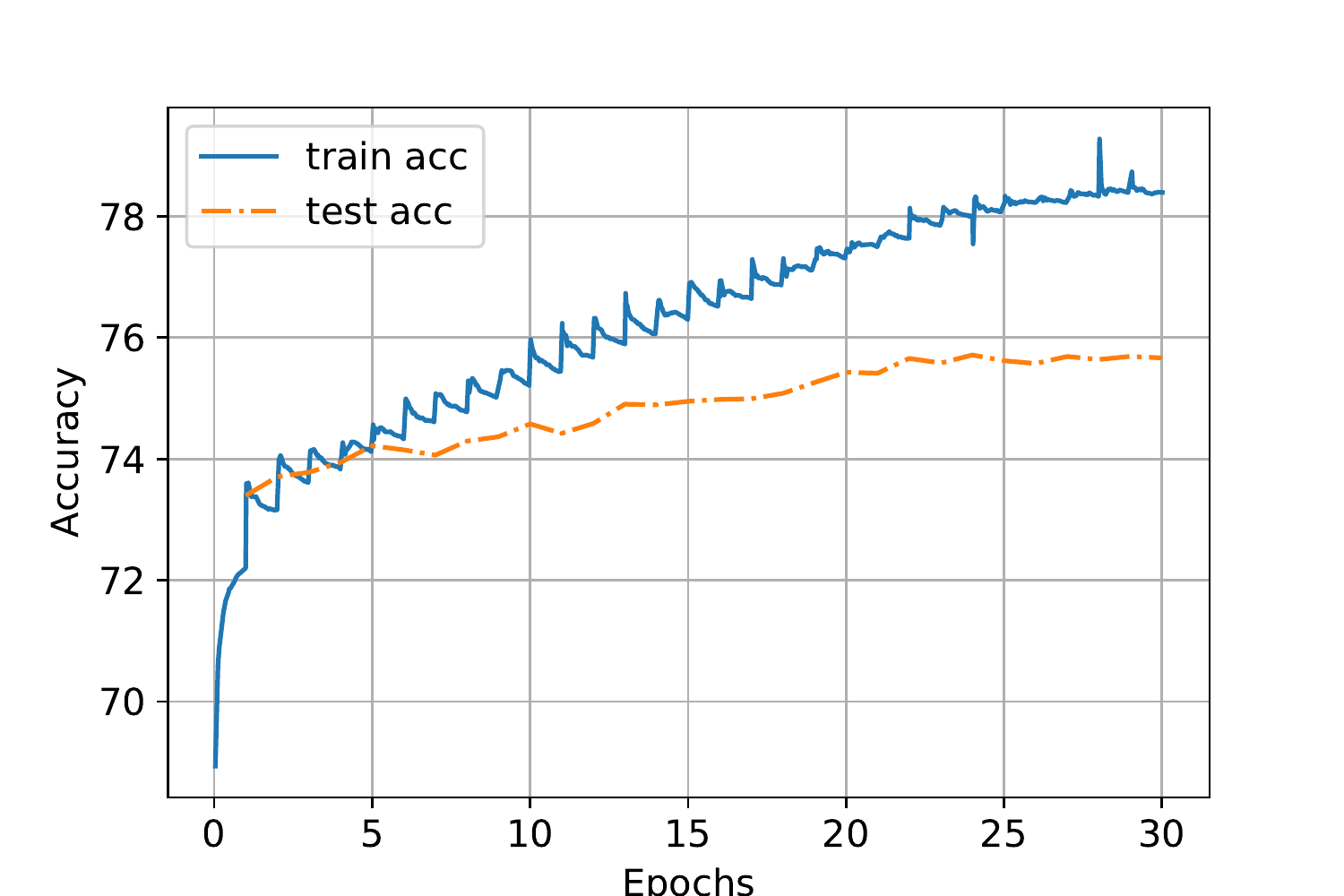}
    \caption{The learning curve of the \textit{(left)} retraining phase and \textit{(right)} fine-tuning phase. The model is ResNet-50 with parameter setting R4.}
    \label{fig:learn_curve}
\end{figure}

\end{document}